\documentclass[sigconf,natbib=false,anonymous=false]{acmart}
\usepackage{multirow}
\usepackage{booktabs,multirow,makecell}
\usepackage{adjustbox}
\usepackage{minted} 
\usepackage{pdfrender}
\usepackage{subcaption}
\usepackage[table]{xcolor}

\setminted{
  breaklines,
  breakanywhere,
  fontsize=\footnotesize,
  linenos,
  tabsize=2
}

\AtBeginDocument{%
  }

\setcopyright{acmlicensed}
\copyrightyear{2026}
\acmYear{2026}
\acmDOI{XXXXXXX.XXXXXXX}
\acmConference[]{SIGIR}{July 20-24, 2026}{MELBOURNE, AUSTRALIA}

\RequirePackage[
  datamodel=acmdatamodel,
  style=acmnumeric,
  ]{biblatex}

\newcommand\method{\texttt{WKGFC}}
\begin{document}

\title{Multi-Sourced, Multi-Agent Evidence Retrieval for Fact-Checking}

\author{Shuzhi Gong}
\orcid{0009-0007-9289-9015}
\affiliation{%
\institution{The University of Melbourne}
\city{Melbourne}
\state{VIC}
\country{Australia}}
\email{shuzhig@student.unimelb.edu.au}

\author{Richard Sinnott}
\orcid{0000-0001-5998-222X}
\affiliation{%
\institution{The University of Melbourne}
\city{Melbourne}
\state{VIC}
\country{Australia}}
\email{rsinnott@unimelb.edu.au}

\author{Jianzhong Qi}
\orcid{0000-0001-6501-9050}
\affiliation{%
  \institution{The University of Melbourne}
  \city{Melbourne}
  \state{VIC}
  \country{Australia}
}
\email{jianzhong.qi@unimelb.edu.au}

\author{Cecile Paris}
\orcid{0000-0003-3816-0176}
\affiliation{%
  \institution{Data61, CSIRO}
  \city{Sydney}
  \state{NSW}
  \country{Australia}}
\email{Cecile.Paris@data61.csiro.au}

\author{Preslav Nakov}
\affiliation{%
  \institution{MBZUAI}
  \city{Abu Dhabi}
  \country{UAE}
}
\email{preslav.nakov@mbzuai.ac.ae}

\author{Zhuohan Xie}
\affiliation{%
  \institution{MBZUAI}
  \city{Abu Dhabi}
  \country{UAE}
}
\email{zhuohan.xie@mbzuai.ac.ae}

\newcommand{\sz}[1]{{\color{blue}{\bf{[SZ:]}} #1}}

\begin{abstract}
Misinformation spreading over the Internet poses a significant threat to both societies and individuals, necessitating robust and scalable fact-checking that relies on retrieving accurate and trustworthy evidence. Previous methods rely on semantic and social-contextual patterns learned from training data, which limits their generalization to new data distributions. Recently, Retrieval Augmented Generation (RAG) based methods have been proposed to utilize the reasoning capability of LLMs with retrieved grounding evidence documents. However, these methods largely rely on textual similarity for evidence retrieval and struggle to retrieve evidence that captures multi-hop semantic relations within rich document contents. These limitations lead to overlooking subtle factual correlations between the evidence and the claims to be fact-checked during evidence retrieval, thus causing inaccurate veracity predictions.

To address these issues, we propose a \textbf{W}eb-enhanced \textbf{K}nowledge \textbf{G}raph retrieval \textbf{F}act-\textbf{C}hecking agentic framework (\textbf{\method}), which exploits authorized open knowledge graph as a core resource of evidence. LLM-enabled retrieval is designed to assess the claims and retrieve the most relevant knowledge subgraphs, forming structured evidence for fact verification. To augment the knowledge graph evidence, we retrieve web contents for completion. The above process is implemented as an automatic Markov Decision Process (MDP): A reasoning LLM agent decides what actions to take according to the current evidence and the claims. To adapt the MDP for fact-checking, we use prompt optimization to fine-tune the agentic LLM. 
Our extensive experiments over datasets in three categories (Wikipedia, websites, and article summaries) show that \method~outperform several advanced state-of-the-art fact-checking methods in balanced accuracy score by over 5\% absolute. These results highlight the effectiveness of knowledge-centric evidence retrieval for fact-checking under open-world settings.

\end{abstract}
\maketitle


\section{Introduction}

The rapid proliferation of misinformation poses a profound threat to public trust, democratic processes, and public health communication~\cite{zhou2020survey}. Efforts to combat misinformation have spanned decades. Traditional studies explored \textit{content semantics}~\cite{hu2021compare,fung2021infosurgeon}, \textit{propagation patterns}~\cite{sun2022ddgcn,bian2020rumor}, and \textit{user activity contexts}~\cite{cui2022meta,gao2022topology} in a pattern-recognition manner. Such approaches depend heavily on training data distributions and often fail to generalize to \textit{unseen domains}, as misinformation topics and linguistic styles evolve over time~\cite{shuzhi2025}. From an information retrieval perspective, this challenge fundamentally concerns how to retrieve reliable and sufficient evidence from large, evolving information sources to support verification.

\begin{figure}[t]
  \centering
  \includegraphics[width=0.43\textwidth]{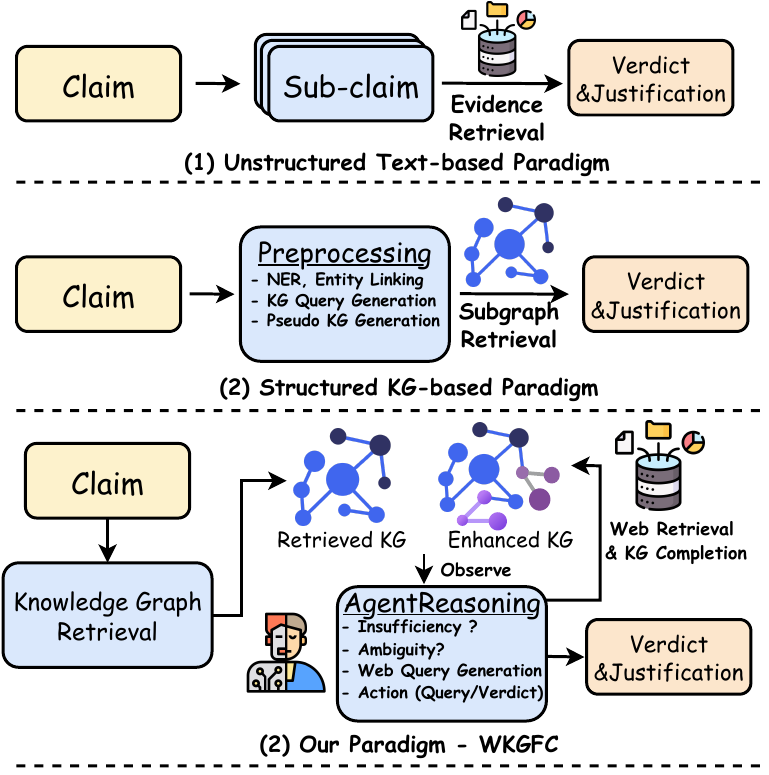}
  \caption{Illustration of different fact-checking paradigms; ours is at the bottom.}
  
  
  \label{fig:illustration_of_now_GFC}
\end{figure}

Recent advances in \textit{Large Language Models} (LLMs) have  transformed automated reasoning and generation capabilities across NLP tasks. \textit{Fact-checking} has emerged as a key frontier where LLMs can expose misinformation and mitigate its societal harms~\cite{chen2024combating}. Yet, LLMs remain limited by their fixed training windows and lack of integration with domain-specific or time-sensitive knowledge, leading to \textit{hallucinated} factual errors. A step forward is the introduction of \textit{Retrieval-Augmented Generation} (RAG)~\cite{nips20-rag}, which enriches LLM reasoning with dynamically retrieved external evidence. The effectiveness of RAG-based fact-checking critically depends on the quality of evidence retrieval, making missing supportive evidence and retrieval failures primary sources of verification errors.

The RAG framework enables a \textit{retrieve--then--verify} paradigm: given a claim to be verified, external knowledge is retrieved and presented to the LLM, which then reasons over both the claim and the retrieved evidence to make a prediction on veracity. Figure~\ref{fig:illustration_of_now_GFC} illustrates this evolution of fact-checking paradigms, highlighting the differences between text-based, graph-based, and our proposed agentic retrieval framework.

A lot of work retrieves \textit{unstructured textual evidence} based on semantic similarity. Systems such as HerO~\cite{yoon-etal-2024-hero} and InFact~\cite{fever2024-infact} established strong RAG baselines by sourcing evidence from large textual knowledge bases. Similar pipelines extend to \textit{open-web retrieval}~\cite{colm2024web,xie2025fire}, i.e., to harness web-scale corpora for evidence gathering. Despite their success, these methods face \textit{semantic ambiguity} and \textit{retrieval uncertainty}: textual similarity does not guarantee factual relevance between claims and retrieved evidence.

Another line of research emphasizes \textit{structured reasoning} based on \textit{knowledge graphs (KGs)}. Early, graph-based methods~\cite{gong2023fake,graph-review} leverage Graph Neural Networks (GNNs) for claim verification. They are constrained by limited representational power and dependence on manually curated \emph{Knowledge Graphs} (KG). More recently, graph-based RAG methods such as GraphRAG~\cite{edge2024graphrag} combine LLM reasoning with KGs, improving factual consistency assessment in complex domains such as medicine~\cite{trumorGPT} and summary verification~\cite{chen2025graphcheck}. Besides, KG question answering frameworks (e.g., ChatKBQA~\cite{luo2024chatkbqa} and SymAgent~\cite{liu2025symagent}) adopt elaborate retrieval mechanisms to extract subgraphs for reasoning, which have been recently adapted to claim verification tasks~\cite{pham2025claimpkg}. While structured methods exhibit stronger factual grounding, they often assume that evidence must exist within the KG, a constraint that does not always hold in open-world settings. Moreover, most of these methods operate on datasets with \textit{gold or domain-specific evidence} (e.g., biomedical or legal corpora~\cite{chen2025graphcheck,sansford2024grapheval}, which limits their generalization capability to broader real-world fact-checking scenarios. These limitations indicate that fact-checking poses a distinct retrieval problem that requires structured reasoning over explicit relations and adaptive retrieval strategies beyond one-shot similarity-based search.

From a retrieval perspective, RAG-based fact-checking remains constrained by three major issues: (i) \textit{Ambiguity in unstructured evidence}, as text-based retrieval often fails to ensure accurate claim--evidence alignment, (ii) \textit{Lack of adaptive retrieval intelligence}, as existing systems lack agentic reasoning to determine \textit{what} to retrieve and \textit{when}, and (iii) \textit{Insufficient integration of structured and unstructured knowledge}, as few frameworks effectively combine KGs with web-scale information to support comprehensive reasoning.

To address these issues, we propose a \textit{multi-sourced, agentic fact-checking framework} that unifies structured KGs and open-web evidence under a single reasoning paradigm. Motivated by the cognitive workflow of human fact-checkers~\cite{amazeen2015revisiting,warren2025show}, our system dynamically determines how, what, or when to retrieve and reason across heterogeneous sources, operating as a reasoning agent as formulated within a Markov Decision Process (MDP). Given a claim, our agent first performs KG retrieval by issuing SPARQL queries over open-resource KGs (e.g., Wikidata or DBpedia) using an \textit{expand-and-prune} search guided by LLM reasoning. The retrieved subgraph serves as the agent’s initial observation. If sufficient evidence is found, the LLM directly performs entity-level reasoning to decide the veracity of an input claim. Otherwise, the agent adaptively selects among several actions, including additional KG expansion or targeted web retrieval, to gather complementary information. This formulation explicitly treats evidence acquisition as a sequential information retrieval process under partial observations.

Unlike closed-world KG question answering, fact-checking operates under an open-world retrieval assumption, where relevant evidence may not exist in structured knowledge bases. To address incomplete or noisy KG coverage, we introduce a web retrieval module that complements KG evidence. Retrieved passages from the open web are filtered through a coarse-to-fine grained pipeline, combining web information retrieval and LLM-based factual consistency evaluation. The surviving passages are then converted into knowledge triplets aligned with the KG schema. The resulting web-enhanced KG fuses factual precision from structured KGs with the contextual breadth of the web, providing a comprehensive evidence space for reasoning.

Building upon the MDP formulation, we further enable policy improvement without model fine-tuning through self-reflection and prompt optimization. After each reasoning trajectory, the agent evaluates its own performance, identifying error types such as missing relations, premature verdicts, or redundant retrievals. These structured reflections populate an experience buffer that guides prompt-level optimization using the TextGrad~\cite{TextGrad} framework. The prompt, treated as a trainable policy parameter, is iteratively refined to maximize a composite reward that considers correctness, evidence coverage, citation quality, and efficiency. It serves as a retrieval decision policy, guiding when to expand, stop, or conclude. This reflection-driven optimization equips the agent with adaptive decision-making capabilities while keeping the base LLM frozen.

We evaluate our proposed Web-enhanced Knowledge Graph Fact-Checking (\method) framework across diverse benchmarks, including Wikipedia-based (FEVER~\cite{fever}, HOVER~\cite{hover}), web-sourced (LIAR-New~\cite{liar-new}, AveriTeC~\cite{averitec-dataset}), and gold-evidence (e.g. SummEval~\cite{fabbri2021summeval}-series) datasets. \method\ reports  consistent improvements over large-scale LLMs and representative baselines, including RAG-based, graph-based, and web-agent methods. These demonstrate the effectiveness of \method\ and its  agentic reasoning and self-reflective optimization.
To the best of our knowledge, our framework provides the first unified agentic solution that:

\begin{itemize}
    \item \textbf{Formulates fact-checking as a multi-source information retrieval MDP process}, enabling the agent to reason and to act across heterogeneous evidence environments.
    \item \textbf{Enables adaptive retrieval over both structured and unstructured knowledge}, by dynamically integrating knowledge graphs and open-web information.
    \item \textbf{Achieves self-improvement through reflection-driven prompt adaptation}, by allowing the agent to refine its decision policy without modifying LLM parameters.
\end{itemize}

By bridging KG reasoning, web retrieval, and LLM decision policies, \method\ advances towards \textit{robust, open-world, and self-adaptive fact-checking}.

\section{Related Work}

\subsection{Fact-Checking Methods}
Fact-checking aims to determine the veracity of a claim. 
Traditional methods relied on linguistic, statistical, and social-contextual features~\cite{zhou2020survey,ying2021fake}, modeling misinformation using textual semantics~\cite{hu2021compare,fung2021infosurgeon}, propagation patterns~\cite{sun2022ddgcn,bian2020rumor}, or user credibility~\cite{cui2022meta,gao2022topology}. From an information retrieval perspective, the core limitation of these approaches lies not in verification itself, but in their inability to reliably retrieve evidence that is both factually relevant and sufficient for decision making.

Although these systems achieved strong in-domain balanced accuracy, they are inherently pattern-recognition models and often fail to generalize to unseen topics, domains, or temporal shifts. This limitation arises from their dependence on dataset-specific language cues and domain distributions.

The advent of LLMs~\cite{gpt4} has transformed fact-checking into a reasoning-driven process~\cite{zheng2025predictions}. RAG-based  frameworks~\cite{shao2023enhancingretrievalaugmentedlargelanguage,trivedi2023interleavingretrievalchainofthoughtreasoning,wei2022chain} integrate external knowledge retrieval with generative reasoning, enabling models to access up-to-date information. Iterative RAGs~\cite{fever2024-infact,colm2024web,xie2025fire,yoon-etal-2024-hero} further enhance factual grounding through multi-step retrieval, but remain limited by semantic similarity retrieval, which often fails to capture deeper logical relations between claims and evidence at retrieval time. Systems such as HerO~\cite{yoon-etal-2024-hero} and InFact~\cite{fever2024-infact} exemplify this line of work, where the retrieved documents are semantically close to the claim but may not correspond factually, leading to false veracity predictions.

Most of these methods rely on unstructured text, lacking mechanisms for multi-hop relational reasoning or handling evidence insufficiency. In contrast, our proposed framework explicitly prioritizes structured retrieval over KGs as the first-stage evidence acquisition process, before invoking open-web retrieval when necessary.

\subsection{LLM Applications over Knowledge Graphs}
KGs provide a structured and interpretable foundation for factual reasoning. Each fact is encoded as a triplet linking entities through semantic relations, allowing for transparent and compositional inference. Early KG-based approaches focused on specialized neural architectures such as GNNs and relational encoders for entity-level classification and reasoning~\cite{Structgpt,reasoningongraph}. These studies demonstrated that structured relational signals can enhance factual precision, especially in domains where relational dependencies are explicit, such as science, medicine, and law.

Recent advances in LLM–KG integration combine symbolic structure with neural reasoning. Systems such as StructGPT~\cite{Structgpt} and RoG~\cite{Rog} allow LLMs to reason over structured inputs through schema-aware prompts or graph-conditioned decoding. Efforts in knowledge-based question answering (KBQA)~\cite{liu2025symagent,luo2024chatkbqa,tog1} highlight how subgraph retrieval and query decomposition improve the grounding of language models. Meanwhile, systmes such as FactKG~\cite{factkg}, ClaimPKG~\cite{pham2025claimpkg}, and GraphCheck~\cite{chen2025graphcheck} explore verification tasks using retrieved or constructed subgraphs, achieving improved factual alignment compared to unstructured baselines.

Though LLM-guided graph reasoning frameworks advance this direction by performing efficient sub-KG discovery, most of these methods assume evidence completeness within the KG or rely on gold-standard data. For example, KBQA tasks assume that the question answers must exist in the KG. FactKG~\cite{factkg} and ClaimPKG~\cite{pham2025claimpkg} assume that the evidence must exist in the KG. Such assumptions rarely hold in real-world scenarios because it reduces fact-checking to a closed-world retrieval problem, which differs fundamentally from real-world verification where evidence availability is uncertain. New claims may refer to evolving or obscure knowledge. Our work bridges this gap by treating KG retrieval as the first stage in a dynamic retrieval decision process, augmenting structured evidence with open-web information only when factual coverage is insufficient.

\subsection{Knowledge Graph Retrieval}
KG retrieval serves as the foundation for structured reasoning and our \method~framework. It plays a central role in domain-specific fact-checking, where evidence must be acquired under incomplete and evolving knowledge assumptions. Classical KGQA pipelines follow a \emph{generate-then-retrieve} paradigm, where a model formulates an explicit query (e.g., SPARQL) to retrieve subgraphs containing relevant entities and relations~\cite{atif2023beamqa,luo2024chatkbqa,yudecaf}. While precise, these methods assume that the KG already contains the correct facts, making them less robust in open-domain or incomplete-resource settings. To improve coverage, modern frameworks adopt an \emph{expand-and-prune} paradigm~\cite{tog2,tog1}. Instead of issuing a single rigid query, the model iteratively expands from seed entities and prunes irrelevant nodes using semantic relevance estimated by an LLM. This strategy ensures multi-hop exploration while maintaining manageable graph size. Beyond traditional retrieval, graph-aware retrieval–generation hybrids combine the efficiency of KG traversal with the flexibility of textual RAG. GraphRAG~\cite{graphrag} introduces graph-context aggregation for summarization and fact verification, demonstrating improved factual grounding in domains such as biomedical text~\cite{trumorGPT}. GraphCheck~\cite{chen2025graphcheck} extends this principle by automatically constructing KGs from long documents, encoding them through GNNs, and reasoning jointly with LLMs. This research shows that graph structure provides an effective scaffold for factual reasoning, especially in multi-hop or long-text contexts.

Recent studies also emphasized the role of intelligent agents and self-improving policies in retrieval workflows. Agentic reasoning frameworks~\cite{liu2025symagent,toolformer,yao2022react,xie2025fire} enable models to plan, retrieve, and reflect iteratively, and to refine their retrieval actions based on feedback. 
Such mechanisms improve efficiency and factual alignment by treating retrieval as a policy optimization problem rather than a static lookup. Our proposed framework inherits these ideas, but specializes them for fact-checking. We formulate KG retrieval, web retrieval, and reasoning as sequential actions within an MDP, allowing an LLM-based agent to decide adaptively when to query, expand, or  conclude with a verdict. 

Overall, existing approaches highlight that fact-checking balanced accuracy is ultimately bounded by evidence retrieval quality, motivating the need for domain-specific IR frameworks that support adaptive, multi-source evidence acquisition.

In summary, while both text-based RAG and KG-based verification contribute valuable perspectives, they face challenges when used in isolation. Textual systems struggle with semantic ambiguity, whereas structured systems lack coverage. Our \method\ framework unifies them through agentic control, ensuring both factual grounding and adaptive coverage in open-world environments.

\section{Preliminaries}
\textbf{Knowledge Graph:} A knowledge Graph (KG) \( \mathcal{G} \) represents facts as triplets of subject entity, relation, and object entity, denoted as \( t = (e, r, e') \), where entities \( e, e' \in \mathcal{E} \) are connected by a relation \( r \in \mathcal{R} \); $r$ can also be referred as $r(e, e')$. In  open-resource KGs, the same relation of a subject entity may correspond to multiple object entities. For example, the \textit{instance\_of} relation in Wikidata describes which topic a subject entity corresponds/belongs to. 
\\[0.1cm]
\textbf{Claim Verification:} Given a claim \( c \), a verification model \( \mathcal{F} \) determines its veracity $v$ as \textit{Supported} or \textit{Refuted} based on an external knowledge base \( \mathcal{K} \), while also providing a justification \( j \) to explain the predicted label. We consider the scenario where \( \mathcal{K} \) is a hybrid of a structured KG \( \mathcal{G} \) and unstructured web information \(\mathcal{K}_{web}\), enabling reasoning over graph knowledge to infer \( v \) and \( j \). Formally, the verification process is defined as: \((v, j) = \mathcal{F}(c, \mathcal{K})\), where \(\mathcal{K} = \mathcal{G} \oplus \mathcal{K}_{web}\).
\\[0.1cm]
\textbf{Markov Decision Process: }
We transform the fact-checking reasoning task on external information into an LLM-based agent task, where the KG and web serve as the environment providing execution feedback rather than merely acting as a knowledge base. The reasoning process can thus be viewed as a multi-step interaction with partial observations from the retrieved evidence. This interactive process can be formalized as a Partially Observable Markov Decision Process (POMDP): $(\mathcal{C}, \mathcal{S}, \mathcal{A}, \mathcal{O}, \mathcal{T})$ with claim space $\mathcal{C}$, state space $\mathcal{S}$, action space $\mathcal{A}$, observation space $\mathcal{O}$, and state transition function $\mathcal{T}: \mathcal{S} \times \mathcal{A} \rightarrow \mathcal{S}$. Note that in our language agent scenario, $\mathcal{C}$, $\mathcal{A}$, and $\mathcal{O}$ are subspaces of the natural language space, and the transition function $\mathcal{T}$ is determined by the environment.

Given a claim to be verified $c \in \mathcal{C}$ and the evidence knowledge base $\mathcal{K}$, the LLM agent generates the action $a_0 \sim \pi_\theta(\cdot \vert c, \mathcal{K}) \in \mathcal{A}$ based on its policy $\pi_\theta$. This action leads to a state transition, and the agent receives execution feedback as observation $o_0 \in \mathcal{O}$. The agent then continues to explore the environment until it is confident to make a final verdict, or the   stop condition is met. The historical trajectory $\mathcal{H}_n$ at step $n$, consisting of a sequence of actions and observations, can be represented as follows:
\begin{equation}\label{history}
\mathcal{H}_n = (q, \mathcal{K}, a_0, o_o, \ldots, a_{n-1},  o_{n-1}) \sim \pi_\theta(\mathcal{H}_n \vert c, \mathcal{K})
\end{equation}

\begin{equation}\label{history1}
\pi_\theta(\mathcal{H}_n \vert c, \mathcal{K}) = \prod_{j=1}^n\pi_\theta(a_j\vert c, \mathcal{K}, a_0, o_0, \ldots,a_{j-1} ,o_{j-1})
\end{equation}
The final reward $r(c, \mathcal{H}_n) \in [0, 1]$ is computed, with 1 indicating a correct prediction.

\begin{figure*}[h]
  \centering
  \includegraphics[width=0.95\textwidth]{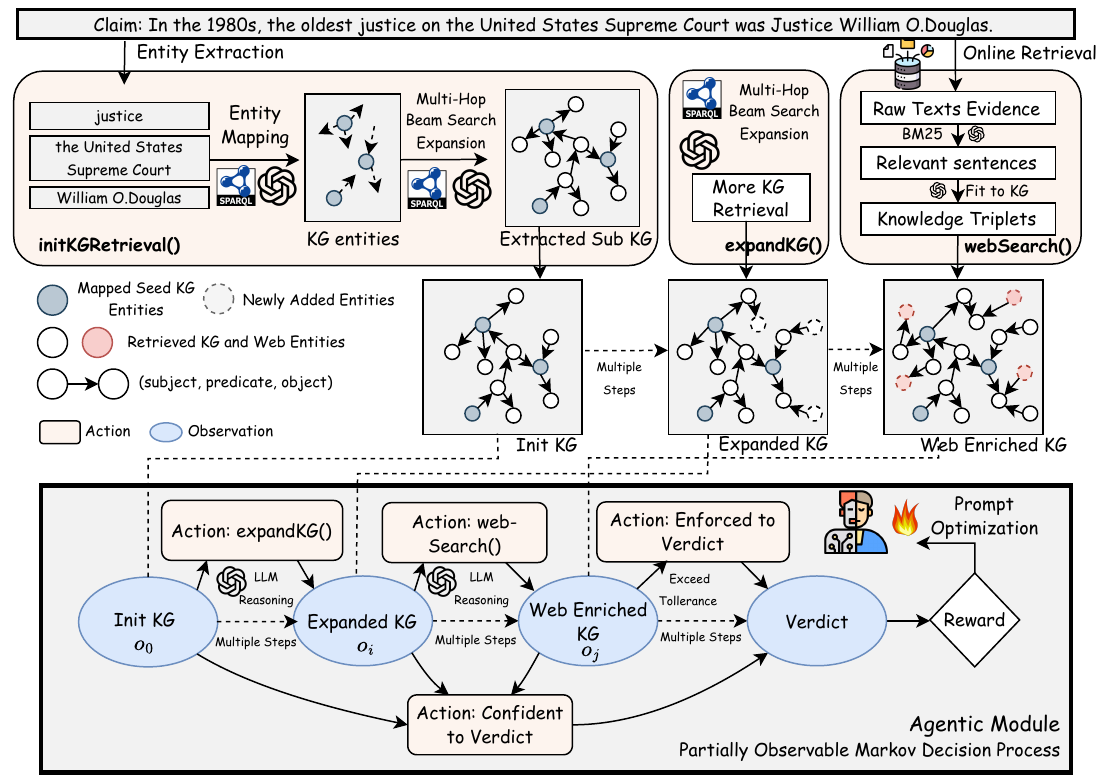}
  \caption{Our agentic framework \method. }
  
  
  \label{fig:wkgfc_framework}
\end{figure*}

\section{Methodology}
Our proposed \method\ framework integrates open-resource KGs and web-based information as external evidence sources, combined with the reasoning ability of LLMs, to achieve automated fact-checking. Given a claim to be verified, the LLM collaborates with pre-defined SPARQL queries to retrieve the most relevant knowledge subgraphs from the KG using an \emph{expand-and-prune} strategy (Section~\ref{Sec:knowledge_graph_retrieval}). 

When the KG evidence is insufficient for verification, a \emph{web retrieval agent} is triggered to collect complementary, up-to-date information from the open web. The retrieved web evidence is then aligned with the KG schema and merged with the KG subgraph to construct an augmented, multi-source knowledge representation. Finally, the LLM reasons over the aggregated evidence to output a veracity verdict and its corresponding justification. The overall architecture is illustrated in Figure~\ref{fig:wkgfc_framework}.

\subsection{Agentic Module}


We model the evidence acquisition and verification process as an agent operating within a POMDP environment. The POMDP formulation explicitly models evidence acquisition as a sequential information retrieval process, where state transitions corresponds to retrieval decisions and observations correspond to retrieved evidence, as shown in Figure~\ref{fig:wkgfc_framework}. 

The first action $a_{0}$ is defined as an initial knowledge graph retrieval based on the claim \texttt{initKGRetrieval(claim)}, and the first observation $o_{0}$ is corresponding initial knowledge subgraph. If this observation is assessed as sufficient, the agent proceeds to make a verdict; otherwise, it selects actions to expand the current KG \texttt{expandKG()}  or complete it by web search \texttt{webSearch()}. 

The action space is defined as function tools to capture distinct retrieval and decision operations. This design allows the agent to seamlessly integrate structured KG data and unstructured web documents, addressing the inherent limitations of LLMs in handling mixed data modalities. We define the following actions:

\begin{itemize}
    \item \texttt{initKGRetrieval(claim)}: Extract entities (i.e., topic entities) from the claim using Named Entity Recognition and entity linking techniques~\cite{entitylinking}. These entities are mapped to KG nodes to retrieve the initial sub-KG, which forms the agent’s first observation.
    
    \item \texttt{expandKG(claim, currentKG, topicEntities)}: Expand the current KG by one additional hop when the evidence is insufficient to make a confident verdict. The expanded KG is then pruned and refined based on relevance to the claim to be verified.
    
    \item \texttt{webSearch(query, currentKG)}: When the KG evidence remains incomplete, the agent reasons and formulates a focused web query guided by the missing information. The retrieved web passages are processed and aligned with the KG schema, bridging structured (triplets) and unstructured (description/annotation) evidence sources.
    
    \item \texttt{verdict(claim, $\mathcal{G}$, $\mathcal{K}_{web}$)}: Perform the final reasoning step to decide the claim’s veracity based on all gathered evidence from the KG ($\mathcal{G}$) and the web ($\mathcal{K}_{web}$). The function outputs both the predicted verdict and the corresponding evidence justification.
\end{itemize} 
This modular design enables the agent to iteratively retrieve, assess, and conclude within a unified decision-making framework, balancing efficiency and performance across heterogeneous evidence sources.

\subsection{Knowledge Graph Retrieval Module}
\label{Sec:knowledge_graph_retrieval}
The objective of KG retrieval in our framework is to acquire a compact, claim-relevant evidence subgraph that is sufficient to support verification. We follows an \emph{expand-and-prune} paradigm~\cite{tog1}, in contrast to the \emph{generate-then-retrieve} strategies commonly used in KGQA systems such as ChatKBQA~\cite{luo2024chatkbqa} and SymAgent~\cite{liu2025symagent}. The latter approaches rely on generating explicit KG queries (e.g., SPARQL) that assume the answer already exists in the KG, which is an unrealistic assumption for open-world fact-checking, where evidence must often be discovered dynamically.

Our retrieval pipeline integrates Wikidata APIs, SPARQL queries, and LLM-based pruning. As shown in Figure~\ref{fig:wkgfc_framework}, the KG retrieval begins with the \emph{initKGRetrieval()}, a Spacy\footnote{\url{https://spacy.io/}}-implemented entity extraction performed on the input claim. Then it is followed by entity mapping to Wikidata nodes done by Wikidata API. The system then executes SPARQL queries to iteratively expand the subgraph while pruning irrelevant branches via beam search guided by the LLM. Here are some implementation details: 

\paragraph{KG Entity Mapping.} The entities are extracted from the claim using SpaCy\footnote{\url{https://pypi.org/project/spacy/}}, then mapped to Wikidata entries through the \texttt{wbsearch entities} API\footnote{\url{https://www.mediawiki.org/wiki/Wikibase/API/}}. These mapped entities serve as seed nodes for the subsequent graph expansion.

\paragraph{Beam Search Expansion.}  
Given the seed entities, the system executes two SPARQL query templates to retrieve both incoming and outgoing relations (Appendix~\ref{Appendix:retrieve_relations}). Because a single relation can connect to multiple entities, the retrieved candidates are ranked using LLM reasoning on semantic relevance to the claim. In each iteration (hop), the top-$k$ most relevant relations and their associated entities are retained, while others are pruned. The process repeats for $N$ hops, resulting in a compact, semantically relevant subgraph serving as the structured evidence base. 

In addition, when the current retrieved knowledge graph is assessed as not sufficient by the Agentic Module, more KG retrieval will be triggered by \emph{expandKG()}, based on more Beam Search Expansion. 
\subsection{Web Retrieval Module}

The web retrieval module complements the KG retrieval by providing additional evidence when the KG coverage is incomplete. However, web information is often noisy and may contain misinformation itself. To mitigate this, web retrieval is only triggered when KG coverage is insufficient and the web evidence is treated as hypothesis-expanding, not authoritative. 

We introduce a two-stage selection and integration pipeline to fuse the web evidence to the retrieved KG. 

\noindent\textbf{Coarse-to-Fine Evidence Filtering.}  
Following standard RAG practices, we first perform coarse-grained retrieval using the Serper\footnote{\url{https://serper.dev}} API to collect top-ranked (BM25\footnote{\url{https://en.wikipedia.org/wiki/Okapi_BM25}}) web documents. Then, a finer-grained LLM filter re-evaluates the candidate passages for factual relevance and claim consistency. Each retained passage is associated with a consistency confidence used downstream.

\begin{table*}[htb]
\centering
\caption{Balanced Accuracy comparison across general and medical fact-checking datasets. The best and the second-best results are shown in \textbf{bold} and \underline{underlined}, respectively, `--' indicates not applicable due to experiment settings, and grey-highlighted columns denote models that use self-retrieved rather than gold evidence.}

\begin{adjustbox}{width=\textwidth}
\renewcommand{\arraystretch}{0.92}
\begin{tabular}{lccccccccc}
\toprule
\textbf{Method} & 
\multicolumn{2}{c}{\textbf{Wikipedia Sourced}} & 
\multicolumn{2}{c}{\textbf{Web Sourced}} &
\multicolumn{4}{c}{\textbf{With Gold Evidence}} &
\makecell{\textbf{Overall}\\\textbf{Avg. (\%)}} \\
\cmidrule(lr){2-3} \cmidrule(lr){4-5} \cmidrule(lr){6-9}
 & \makecell{FEVER} & \makecell{HOVER} & LIAR-New & AveriTec 
 & SummEval & AFact-CNN & PubHealth & Avg.(\%) \\
\midrule
\multicolumn{9}{l}{\textit{\textbf{Large-scale LLMs}}} \\
Gemini-2.5-flash & 54.8 & 62.2 & 59.1 & 61.0 & 69.9 & 64.6 & 70.8 & 68.4 & 63.2 \\
GPT-4 & 51.4 & 62.5 & 58.6 & 60.2 & 69.7 & 60.7 & 73.2 & 67.9 & 62.3\\
GPT-4o & 55.3 & 63.8 & 60.1 & 61.5 & \underline{76.3} & \underline{66.8} & 67.0 & \underline{70.0} & 64.4 \\
Claude 3.5-Sonnet & 57.1 & 64.4 & 61.3 & 63.0 & \textbf{77.3} & \textbf{68.8} & 73.8 & \textbf{73.3} & 66.5 \\
DeepSeek-V3 67B & 53.5 & 61.7 & 58.8 & 60.5 & 68.3 & 63.2 & 72.9 & 68.1 & 62.7 \\

\multicolumn{9}{l}{\textit{\textbf{Small-scale LLMs}}} \\
Llama3 8B~\cite{dubey2024llama3} & 48.2 & 55.1 & 50.3 & 54.7 & 51.7 & 51.3 & 70.7 & 57.9 & 54.6 \\
Qwen2.5 7B~\cite{yang2025qwen3} & 50.1 & 57.6 & 52.9 & 56.3 & 58.5 & 45.3 & 59.1 & 54.3 & 54.3 \\
Llama3.3 70B~\cite{dubey2024llama3} & 57.4 & 61.2 & 60.5 & 62.1 & 57.6 & 53.5 & \textbf{76.9} & 62.7 & 61.3\\
Qwen2.5 72B~\cite{yang2025qwen3} & 58.1 & 62.0 & 59.7 & 61.4 & 53.4 & 49.9 & \underline{76.7} & 60.0 & 60.2\\

\multicolumn{9}{l}{\textit{\textbf{Specialized Methods}}} \\
HerO~\cite{yoon-etal-2024-hero} & 67.5 & 58.1 & 70.2 & 71.4 & 65.8 & 62.4 & 70.3 & 66.2 & 66.5\\
GraphRAG (GPT-4o)~\cite{graphrag} & - & - & - & - & 68.2 & 60.4 & 74.2 & 67.6 & - \\
GraphCheck~\cite{chen2025graphcheck} & - & - & - & - & 66.3 & 61.9 & 70.9 & 66.4 & -\\
FIRE~\cite{xie2025fire} & \underline{90.6} & \underline{67.0} & \underline{72.8} & \underline{72.8} & 
\cellcolor{gray!25}58.2 & \cellcolor{gray!25}56.9 & \cellcolor{gray!25}64.1 & \cellcolor{gray!25}59.7 & \underline{68.9} \\
\method (Ours) & \textbf{91.9} & \textbf{72.8} & \textbf{81.3} & \textbf{73.2} & 
\cellcolor{gray!25}69.8 & \cellcolor{gray!25}58.4 & \cellcolor{gray!25}72.8 & \cellcolor{gray!25}67.0 & \textbf{74.3}\\
\bottomrule
\end{tabular}
\end{adjustbox}
\label{tab:graphcheck-performance}
\end{table*}

\noindent\textbf{Web-to-KG Integration.}  
To unify the structured and the unstructured evidence, selected web sentences are converted into knowledge triplets and annotations to be aligned with the KG schema. If a triplet already exists in the KG, it is skipped; if it enriches an existing entity (e.g.,~by providing explanatory details), it is appended as a textual annotation. This controlled fusion constructs an \emph{augmented knowledge graph} that captures both factual precision from structured KGs and contextual breadth from the web. KG-oriented facts are treated as higher-precision anchors and web-derived evidence is treated as expansions. 

By maintaining the agentic reasoning loop across KG and web retrievals, our framework enables dynamic, multi-source evidence synthesis, thus allowing the LLM agent to reason under partial observations and iteratively approach a reliable verdict.

\subsection{Policy Improvement via Self-Reflection and Prompt Optimization}

Within the POMDP formulation, the agent’s policy $\pi_{\theta}$ determines which retrieval or decision action to take given the current observation, which consists of the accumulated evidence state. In our framework, this policy is instantiated by the reasoning behavior of the agentic module. While the intrinsic parameters of the LLM remain frozen during inference, the policy can be refined at the behavioral level by modifying the prompt that governs how the LLM selects actions. This design enables policy improvement without gradient-based model updates, allowing the agent to adapt its retrieval behavior through experience while remaining fully parameter-free.

We adopt a two-stage policy improvement scheme inspired by self-reflective LLM prompt optimization~\cite{khattab2024dspy,TextGrad}, consisting of structured self-reflection followed by prompt-level optimization. Importantly, this process targets retrieval coordination and stopping decisions. 

\noindent\textbf{Self-Reflection.}
For each claim verification episode, the agent produces a complete decision trajectory $H_n$, represented as a sequence of retrieval and reasoning steps (e.g., \texttt{InitialKG} $\rightarrow$ \texttt{Reasoning} $\rightarrow$ \texttt{EnhancedKG} $\rightarrow$ \texttt{Reasoning} $\rightarrow$ \texttt{Web-EnhancedKG} $\rightarrow$ \texttt{Verdict}). After generating a final prediction, the agent evaluates the outcome and produces a structured self-critique that attributes success or failure to specific decision patterns. These critiques focus on retrieval-related error types, including insufficient evidence coverage, premature termination of retrieval, redundant expansion steps, or mismanagement of contradictory evidence.

Each time step in the trajectory is recorded in an experience buffer
$\mathcal{D}=\{(s_t, a_t, o_t, r_t, \text{critique}_t)\}$,
where $s_t$ denotes the belief state, $a_t$ the selected action, $o_t$ the resulting observation, $r_t$ a scalar reward signal, and $\text{critique}_t$ the corresponding textual feedback. This buffer captures how retrieval decisions influence verification outcomes across diverse evidence environments, forming the basis for subsequent policy refinement.

\noindent\textbf{Prompt Optimization.}
Rather than updating model weights, we treat the agent’s prompts as trainable policy descriptors $\theta$ and optimize it directly over the experience buffer using the TextGrad framework~\cite{TextGrad}. TextGrad interprets textual critiques as gradient-like constructive signals, enabling iterative refinement of the prompt to better align action selection with effective retrieval behavior. In this context, prompt optimization serves as an outer-loop policy improvement mechanism that shapes how the agent prioritizes retrieval depth, source selection, and stopping conditions.

Each trajectory is associated with a composite reward that reflects (i) prediction correctness, (ii) evidence sufficiency, measured by whether retrieved information adequately supports or refutes the claim, and (iii) retrieval efficiency, penalizing unnecessary or redundant retrieval actions. Contradictory evidence between KG and web sources is not treated as noise to be eliminated; instead, failures arising from improper handling of such contradictions are explicitly surfaced in self-critiques and incorporated into the optimization signal.

Prompt optimization is conducted offline using an experience buffer constructed from 150 non-test samples drawn from the FEVER~\cite{fever} and HOVER~\cite{hover} datasets. Optimization is performed for a fixed budget of 20 epochs over 100 training samples' trajectories, with validation conducted on a held-out set of 50 trajectories. The final prompts $\theta^\star$ are selected based on validation performance, ensuring stability and mitigating overfitting. Throughout this process, the base LLM parameters remain unchanged.

The resulting policy $\pi_{\theta^\star}$ exhibits improved retrieval coordination, more reliable stopping decisions, and better handling of evidence conflicts, leading to higher verification accuracy and reduced retrieval cost. Empirical comparisons before and after prompt optimization are reported in Table~2, demonstrating that policy refinement at the prompt level yields consistent performance gains without sacrificing interpretability or scalability.

\section{Experiment and Evaluation}

\subsection{Datasets and Statistics}

We evaluate our framework on diverse fact-checking benchmarks covering multiple evidence sources. The datasets are selected to represent three distinct categories: (i) Wikipedia-based evidence (FEVER, HOVER), which are extracted from the Wikipedia, (ii) Web-sourced evidence (LIAR-New, AveriTeC), which are drawn from web resources, and (iii) Provided textual evidence (SummEval, AggreFact-CNN, PubHealth), which are summarized from gold evidential paragraphs. This design ensures comprehensive evaluation across structured, unstructured, and provided-evidence settings. 

\paragraph{FEVER}~\cite{thorne-etal-2018-fever} contains human-written claims labeled as \textit{SUPPORTED}, \textit{REFUTED}, or \textit{NOT ENOUGH INFO (NEI)}, each paired with Wikipedia evidence, serving as a standard single-hop fact verification benchmark.  

\paragraph{HOVER}~\cite{jiang-etal-2020-hover} extends to multi-hop (i.e., 2 to 4) reasoning across up to four Wikipedia articles, merging \textit{REFUTED} and \textit{NEI} into \textit{NOT-SUPPORTED} for more challenging evidence aggregation.  

\paragraph{LIAR-New}~\cite{liar-new} updates PolitiFact claims beyond 2021 to reduce pretraining leakage, including multilingual claims and rebalanced truth distributions for open-world settings.  

\paragraph{AveriTeC}~\cite{averitec-dataset} offers large-scale, web-sourced claims paired with retrieved evidence documents, reflecting real-world verification scenarios where information is dispersed online.

\paragraph{SummEval-Series}~\cite{chen2025graphcheck} covers factual consistency in long-form text (e.g.,~summarization, biomedical claims), testing models' ability to verify complex summaries against their source passages.\\

\noindent\textbf{Baseline Setup. }  
To ensure fair comparison across all datasets, we standardize the labels into two classes: \textit{Factual Supported} and \textit{Factual Refuted}. The samples lacking verifiable evidence are removed, and partially correct or ambiguous cases are merged into \textit{Factual Refuted}. This unified benchmark allows consistent evaluation across datasets and evidence types. Detailed statistics and information are shown in Table~\ref{tab:dataset_stats}. 

\begin{table}[tbh]
\small
\centering
\caption{The statistic of Datasets. (Neg Rate represents the ratio of evidential refuted samples.)}
\scalebox{0.9}{
\begin{tabular}{lccc}
\toprule

\textbf{Datasets} & \textbf{Size}  & \textbf{Have Evidence} & \textbf{Neg Rate (\%)} \\
\midrule
FEVER & 1,332 & No & 50.0 \\
HOVER & 1,332  & No & 50.0 \\
LIAR-New & 200 & No & 76.0 \\
AveriTeC & 465 & No & 73.7\\
SummEval & 1,600 & Yes & 18.4\\
AggreFact-CNN & 558 & Yes & 10.2 \\
PubHealth & 1,232 & No & 51.3\\

\bottomrule       
\end{tabular}}
\label{tab:dataset_stats1}
\end{table}

\subsection{Baselines}
We compare our proposed framework \method\ against three families of systems listed in Table~\ref{tab:graphcheck-performance}. All methods are evaluated under a unified protocol with standardized binary labels, ensuring a fair comparison across heterogeneous datasets and evidence types. 

\noindent\textbf{General-purpose LLMs (large-scale).}  
We include closed proprietary and open-weight models as strong general LLM baselines: GPT-4, GPT-4o, Claude~3.5-Sonnet, Gemini-2.5-flash, and DeepSeek-V3~67B.  
These models are prompted as generic veracity judges under the same evaluation setup. We report balanced accuracy across all evidence categories to benchmark the reasoning ability of large LLMs without external retrieval.

\noindent\textbf{General-purpose LLMs (small-scale).}  
To contextualize performance versus model scale, we also evaluate open-weight models such as Llama3~8B, Qwen2.5~7B, Llama3.3~70B, and Qwen2.5~72B using the same fact-checking prompt configuration. These models demonstrate the lower bound of zero-shot factual reasoning without retrieval augmentation.

\noindent\textbf{Specialized fact-checking and agentic methods.}  
HerO~\cite{yoon-etal-2024-hero} represents a strong \textit{RAG-based} pipeline with coarse-to-fine document ranking over large evidence databases. In our setup, it consumes dynamically retrieved web passages as its evidence source for real-time verification.  
FIRE~\cite{xie2025fire} is an \textit{iterative retrieval-and-verification} framework that alternates retrieval and reasoning to refine factual grounding, serving as another text-based comparison.  
GraphRAG (GPT-4o)~\cite{graphrag} integrates structured graph context aggregation with LLM reasoning and is evaluated on \textit{gold-evidence} datasets where structured inputs are available.  
GraphCheck~\cite{chen2025graphcheck} constructs knowledge graphs from provided evidence and performs \textit{GNN-augmented} reasoning. 

\noindent\textbf{Evaluation fairness.}  
Across all methods that requires retrieval (e.g.,~HerO, FIRE) or gold evidence (e.g.,~GraphRAG, GraphCheck), we align the input evidence scope or limit evaluation to appropriate dataset categories. This ensures that performance differences stem from reasoning strategy rather than evidence accessibility. For example, for large-scale and small-scale LLMs, they are asked to predict the veracity of claims directly on Wikipedia and web sourced datasets, and are able to refer to the gold evidence on SummEval-series dataset. GraphRAG and GraphCheck are only evaluated on SummEval-series because they require gold evidence to perform graph-based retrieval. For open-sourced retrieval enabled methods (e.g., FIRE and \method), we do not use the SummEval's provided evidence but let them to retrieve evidence by themselves, as indicated in Table~\ref{tab:graphcheck-performance}.

\subsection{Experimental Settings}
Fact-checking research is often fragmented, with methods relying on different types of evidence sources. To establish a fair comparison, we design a unified experimental protocol across datasets that differ significantly in content. Specifically, FEVER, HOVER, and LIAR-New provide only claims with veracity labels; AveriTeC offers a large corpus of candidate evidence; and the SummEval-series datasets supply concise, curated evidence paragraphs for each claim.

This heterogeneity introduces challenges when applying to existing baselines. For instance, GraphCheck~\cite{chen2025graphcheck} is tailored to the concise evidence provided in SummEval and is thus limited when confronted with open-domain claims lacking such gold-standard evidence. To preserve these methods' original implementation, we limit the evaluation of GraphCheck and GraphRAG to the given textual evidence category. For methods requiring GPU computing (e.g.,~HerO, Llama3~8B), we use 4 NVIDIA A100 80GB in our experiments. For methods requiring LLM APIs (e.g.,~FIRE), we use the APIs claimed in their original papers. For our \method, we experiment on top of the OpenAI GPT-4o API. 
Note that, for FIRE and \method, we do not use the gold evidence, but the provided retrieved evidence, as highlighted in Table~\ref{tab:graphcheck-performance}.

\subsection{Performance Analysis}

Table \ref{tab:graphcheck-performance} compares our proposed Web-enhanced Knowledge Graph Fact-Checking (\method) framework to both large-scale LLMs and specialized fact-checking baselines across three evidence categories: Wikipedia-sourced, Web-sourced, and Gold-evidence benchmarks.
We can see that \method~achieved the highest overall balanced accuracy (74.3\%), outperforming the best-performing baseline (FIRE, 68.9\%) by +5.4 points and exceeding all large-scale LLMs such as Claude 3.5 Sonnet (66.5\%) and GPT-4o (64.4\%) despite operating without any task-specific fine-tuning.

\noindent\textbf{Wikipedia-Sourced Datasets.}
On FEVER and HOVER, \method~yields 91.9\% and 72.8\% balanced accuracy, surpassing text-based RAG systems (e.g.,~HerO 58.1\%) and FIRE (90.6\% on FEVER, but 67.0\% on HOVER).
This indicates that integrating knowledge-graph retrieval prior to web reasoning yields stronger factual grounding and better multi-hop inference.
The gain on HOVER further highlights \method's ability to perform compositional reasoning across Wikipedia articles, an ability that similarity-based retrievers often do not have.

\noindent\textbf{Web-Sourced Datasets.}
For LIAR-New and AveriTeC, \method~achieves 81.3\% and 73.2\%, outperforming HerO (70.2 / 71.4) and FIRE (72.8 / 72.8).
These results demonstrate that agentic retrieval policies improve veracity prediction in open-domain contexts where relevant information is distributed across heterogeneous web pages.
Unlike baselines that rely on fixed retrieval rounds, \method's adaptive reasoning dynamically decides whether to expand the KG or to trigger web retrieval, leading to higher evidence coverage with fewer redundant calls.

\noindent\textbf{Gold-Evidence Datasets.}
Under the gold-evidence group (SummEval, AFact-CNN, PubHealth), \method~maintains strong balanced accuracy (67.0\% avg.), which is on par with specialized graph-based models (GraphRAG 67.6, GraphCheck 66.4).
Although those baselines rely on gold annotations, which are unavailable in open settings, \method~ achieves comparable or better results using only retrieved evidence, validating the reliability of its knowledge-grounded reasoning pipeline.
This robustness across both short- and long-form factuality tasks underscores the generality of our proposed framework.

\noindent\textbf{Overall Trends.}
\method~exhibits a consistent upward trend across all evidence categories, showing both high factual precision (Wikipedia tasks) and adaptability (web tasks).
Its ability to unify structured and unstructured knowledge under a single agentic workflow results in a balanced performance profile, rather than excelling only in one domain as most baselines do.
The strong results against 70 B–100 B-parameter LLMs also emphasize the effectiveness of retrieval-driven reasoning and self-reflective policy optimization over sheer model scale.
Together, these results substantiate \method{} as a robust, data-efficient, and interpretable fact-checking system suitable for real-world multi-source scenarios.

\subsection{Ablation Experiment}
\label{Sec:ablation_experiment}
\textbf{Variant 1. KG retrieval only}: In the first variant, we only utilize the evidence from Wikidata to make a verdict. 

\noindent\textbf{Variant 2. Web retrieval only}: Only web source is retrieved.

\noindent\textbf{Variant 3. KG retrieval + agentic module}: This variant enables agentic reasoning, thus allowing more flexible KG retrieval, the agentic module is also optimized through prompt tuning. 

\noindent\textbf{Variant 4: KG retrieval + Web retrieval w/o agentic module}: This variant enables web retrieval, but limits the agentic actions. The web queries are based on the questions but not generated from existing KG evidence. 

\noindent\textbf{Variant 5: \method~w/o agent optimization}: This variant enables all KG, web, and agentic modules, but does not optimize the prompts.


\begin{table}[h]
    \centering
    \caption{Ablation experiments for the \method{} (balanced accuracy in \%).}
    \resizebox{\linewidth}{!}{
    \begin{tabular}{c|cccc|cc|cc}
    \toprule
    & \multirow{2}{*}{\textbf{KG}} 
    & \multirow{2}{*}{\textbf{Web}} 
    & \multirow{2}{*}{\textbf{Agent}} 
    & \multirow{2}{*}{\textbf{Optim}} 
    & \multicolumn{2}{c|}{\textbf{Wiki-sourced}} 
    & \multicolumn{2}{c}{\textbf{Web-sourced}} \\
    \cmidrule(){6-7} \cmidrule(){8-9}
    & & & & & FEVER & HOVER & LIAR-New & AveriTec \\
    \midrule
    \textbf{Variants} 
        & \checkmark & - &- & - 
        & 72.3 & 56.3 & 61.8 & 63.4 \\
        & - & \checkmark & - & - 
        & 79.6 & 66.2 & 68.9  & 72.4 \\
        & \checkmark & \checkmark & - & -
        & 85.4 & 66.6 & 69.0 & 69.4 \\
        & \checkmark & - & \checkmark & \checkmark
        & 76.5 & 64.2 & 65.7 & 68.5 \\
        & \checkmark & \checkmark & \checkmark & - 
        & 89.1 & 67.9 & 69.8 & 71.0 \\
    \midrule
    \method 
        & \checkmark & \checkmark & \checkmark & \checkmark 
        & \textbf{91.9} & \textbf{69.8} & \textbf{70.8} & \textbf{73.2} \\
    \bottomrule
    \end{tabular}}
    \label{Tab:ablation}
\end{table}

Table \ref{Tab:ablation} examines the contributions of each module (knowledge-graph retrieval, web retrieval, and agentic reasoning) on four representative datasets.
When using only KG retrieval, the performance drops notably (e.g.,~FEVER 72.3, AveriTeC 63.4), reflecting limited evidence coverage.
Adding agentic actions on top of KG retrieval improves the balanced accuracy to 76.5–68.5, confirming the benefit of dynamic decision-making in when to retrieve and when to reason.
Finally, enabling both web retrieval and agentic control (full \method) yields the best results (e.g., FEVER 91.9, AveriTeC 73.2), suggesting complementarity between structured KG evidence and unstructured web information.
These ablations verify that the performance gains arise not merely from larger evidence pools, but from the policy-guided integration of heterogeneous sources.

\subsection{KG Retrieval Parameter Impact}
In the KG retrieval module, several parameters, specifically the beam size $k$ and the number of hops $N$, control the exploration scope of the beam search. It is therefore worth investigating \textbf{whether a broader retrieval scope (i.e., larger beam sizes and deeper hops) consistently leads to better performance}. To study this, we analyze the impact of these two parameters on fact-checking balanced accuracy on FEVER and HOVER. To ensure a clean comparison and to avoid interference from web information, we conduct experiments under \textit{Variant 1} from the ablation study (Section~\ref{Sec:ablation_experiment}).

\begin{figure}[tbh]
    \centering
    \begin{subfigure}[t]{0.49\linewidth}
        \centering
        \includegraphics[width=\linewidth]{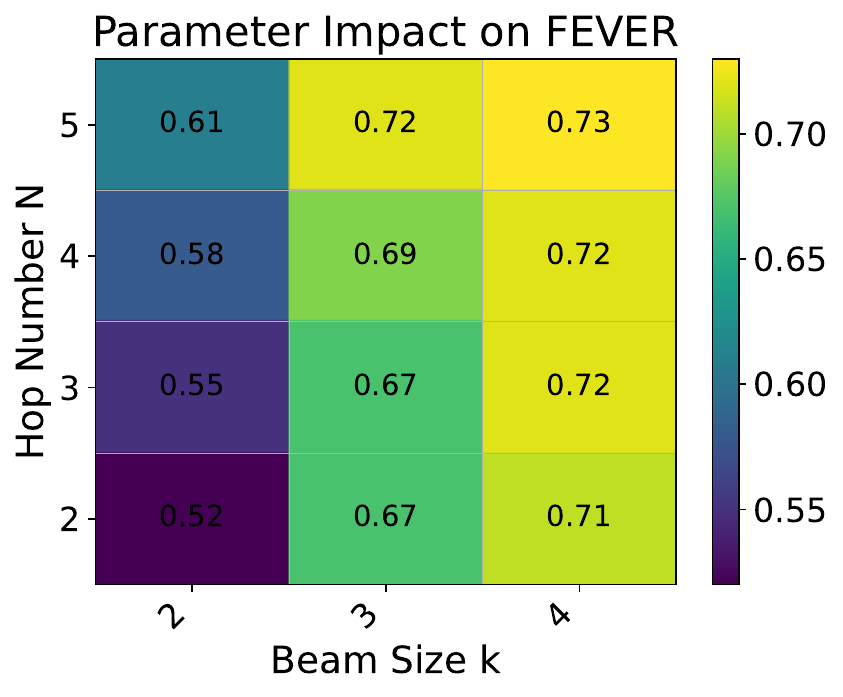}
        \label{fig:heatmap_a}
    \end{subfigure}
    \hfill
    \begin{subfigure}[t]{0.49\linewidth}
        \centering
        \includegraphics[width=\linewidth]{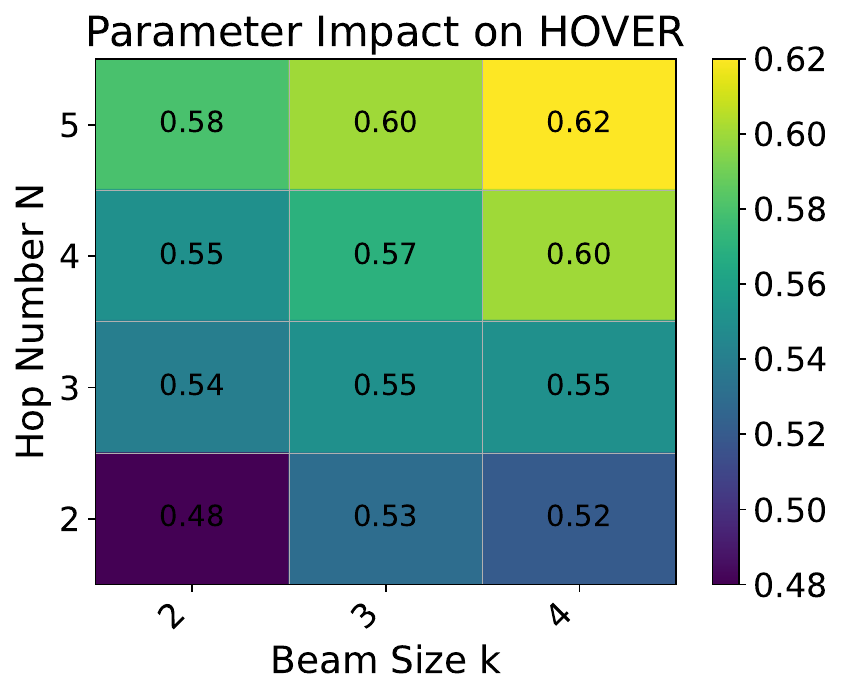}

        \label{fig:heatmap_b}
    \end{subfigure}
    \caption{Parameter study on FEVER and HOVER (balanced accuracy in \%).}
    \label{fig:heatmap_comparison}
\end{figure}

We vary the beam size $k$ from 3 to 5 and the number of hopes $N$ from 2 to 5. The experimental results on FEVER and HOVER are presented in Fig.~\ref{fig:heatmap_comparison}.

Overall, the results indicate that a broader search scope generally improves predictive balanced accuracy. On the single-hop FEVER dataset, performance saturates quickly when $k=4$, even with a small number of hops. In contrast, the multi-hop HOVER dataset benefits from deeper traversal (at least four hops) to achieve strong balanced accuracy.

However, these improvements come at a computational cost. A broader search requires more SPARQL queries and LLM invocations. Under the \textit{Variant 1} configuration, after the entities are extracted and mapped to the KG, the retrieval process can require up to $k \times N$ SPARQL queries (at each hop, k entities to expand) to construct the subgraph and up to $(N + k \times N + 1)$ LLM calls ($N$ calls for pruning at each hop, $k \times N$ calls for entity expansion pruning, $1$ call for final verdict) to reach a verdict. To balance performance and efficiency, we set $k=4$ and $N=4$ in our main experiments.

\subsection{Error Analysis}
\label{error_analysis}

To gain an insight into the \method's performance, we performed a thorough error analysis on the data samples from FEVER, HOVER, LIAR-New, AveriTec and AggreFact-CNN that \method~failed (incorrect predictions). 
We define three types of observations for this analysis: 

(i) \emph{Insufficient KG}: An insufficient KG occurs when the agentic module requests a web retrieval even after the retrieved KG is enhanced. This indicates that the information from the KG is not sufficient to support a prediction. 
(ii) \emph{Exceeding the Maximum Steps}: Exceeding the maximum steps occurs when the agentic module performs all allowed retrieval steps, but is still not confident to make a final verdict. In this case, we use a special prompt to enforce the agent to give a verdict based on the existing evidence.
(iii) \emph{Over-confidence}: In contrast to \emph{Exceeding the Maximum Steps}, over-confidence occurs when the agent does not perform any additional retrieval actions and predicts wrongly based on the initial KG only. 

\begin{table}[t]
\small
\centering
\caption{The statistic of Error Rate (\%) of \method.}
\scalebox{0.9}{
\begin{tabular}{lcccccc}
\toprule

\textbf{Error} & \textbf{FEVER}  & \textbf{HOVER} & \textbf{LIAR-New} & \textbf{AveriTeC} & \textbf{AFact-CNN} &  \\
\midrule
IK & 29.6 & 74.6 & 83.3 & 75.0  & 98.3  \\
EM & 4.6  & 74.1 & 60.0 & 72.5  & 98.0    \\
OC & 64.8 & 19.8 & 16.7 & 15.0  & 1.3        \\

\bottomrule       
\end{tabular}}
\label{tab:error_analysis}
\end{table}

Note that these three observations do not cover all failure cases, as some failures arise from execution errors, and there is one scenario in which the agent requests additional KG hops without requesting a web search (i.e., it is neither overconfident nor limited by insufficient KG information). The results reveal that insufficient KG coverage dominates the errors in complex and less Wiki-related datasets, which motivates the need for better KG construction and retrieval. In contrast, exceeding-step errors are more frequent in multi-hop reasoning tasks such as HOVER (74.1\%), where the agent exhausts retrieval attempts without gaining confidence, which indicates challenges in balancing retrieval depth and decision thresholds. Finally, over-confidence errors are more pronounced in single-hop datasets such as FEVER (64.8\%), reflecting cases where the agent prematurely concludes without sufficient evidence. Overall, the error distribution highlights that improving KG completeness and adaptive stopping strategies is crucial for enhancing \method's robustness and decision reliability.

\section{Conclusion and Future Work}

We presented \method, a unified agentic framework that formulates fact-checking as a domain-specific information retrieval problem under open-world and partially observable conditions. By modeling evidence acquisition and verification as a Partially Observable Markov Decision Process (POMDP), \method\ enables an LLM-based agent to adaptively control retrieval and stopping decisions across structured knowledge graphs and unstructured web sources. The framework integrates expand-and-prune KG retrieval, coarse-to-fine web retrieval, and self-reflective prompt-based policy improvement, allowing retrieval behaviors to be refined without modifying model parameters. Extensive experiments on Wikipedia-based, web-sourced, and gold-evidence benchmarks demonstrate that \method\ consistently outperforms large-scale LLMs and representative baselines such as HerO, FIRE, and GraphCheck, particularly in open-world scenarios where evidence is incomplete or unavailable a priori.

In future work, we plan to extend this retrieval-centric framework toward collaborative multi-agent verification, enabling coordinated evidence acquisition, credibility assessment, and causal reasoning for large-scale misinformation mitigation.

\printbibliography

\end{document}